\documentclass[conference]{IEEEtran}
\IEEEoverridecommandlockouts

\usepackage[utf8]{inputenc}
\usepackage[T1]{fontenc}
\usepackage{cite}
\usepackage{amsmath,amssymb,amsfonts}
\usepackage{graphicx}
\usepackage{booktabs}
\usepackage[table]{xcolor}
\usepackage{array}
\usepackage{tabularx}
\usepackage{algorithm}
\usepackage{algpseudocode}
\usepackage{url}
\usepackage{cleveref}

\newcommand{\methodname}{LiveVLN}
\newcommand{\stopaction}{\texttt{STOP}}

\begin{document}

\title{LiveVLN: Breaking the Stop-and-Go Loop in Vision-Language Navigation}

\author{
\IEEEauthorblockN{Author Name Placeholder}
\IEEEauthorblockA{
\textit{Affiliation Placeholder}\\
City, Country\\
email@placeholder.com
}
}

\author{
\IEEEauthorblockN{
Xiangchen Wang\textsuperscript{1,*},
Weiye Zhu\textsuperscript{1,*},
Teng Wang\textsuperscript{1},
TianTian Geng\textsuperscript{1}, \\
Zekai Zhang\textsuperscript{1},
Zhiyuan Qi\textsuperscript{1},
Jinyu Yang\textsuperscript{2,\textdagger},
Feng Zheng\textsuperscript{1,3,\textdagger}
}
\IEEEauthorblockA{
\textsuperscript{*} Equal contribution. \textsuperscript{\textdagger} Corresponding author.\\
\textsuperscript{1} Southern University of Science and Technology \,
\textsuperscript{2} Harbin Institute of Technology, Shenzhen
\,
\textsuperscript{3} Spatialtemporal AI
}
}
\maketitle

\begin{abstract}
Recent navigation systems achieve strong benchmark results, yet real-world deployment often remains visibly stop-and-go. This bottleneck arises because the sense-inference-execution loop is still blocking: after each new observation, the controller must wait for sensing, transmission, and inference before motion can continue. Reducing action-generation cost alone therefore does not remove redundant waiting. 
To address this issue, we present \methodname, a training-free framework for more continuous embodied navigation by augmenting pretrained VLM navigators with multi-step action continuation.
Instead of pausing for each full sense-and-inference round, \methodname\ overlaps execution with the processing of newly arrived observations, allowing refreshed future actions to be handed off before the current executable prefix is exhausted.
This design keeps actions continuously available during motion, reducing idle waiting and enabling smoother online execution. 
The framework operates at runtime and can be integrated with compatible pretrained VLM navigators.
Across R2R and RxR, \methodname\ preserves benchmark performance while reducing waiting time and improving action availability. 
In real-world deployments, it cuts average episode waiting time by up to $77.7\%$ and shortens wall-clock episode time by $12.6\%$ on StreamVLN and $19.6\%$ on NaVIDA, yielding more coherent execution during deployment. Code is available at \url{https://github.com/NIneeeeeem/LiveVLN}.
\end{abstract}




\section{Introduction}

Vision-Language Navigation (VLN) studies how embodied agents follow language instructions from egocentric visual observations~\cite{Anderson2018R2R,Ku2020RxR,Hong2021VLNBERT,Chen2022DUET}. Classical VLN systems built on stronger cross-modal pretraining and Transformer-based reasoning, such as VLN-BERT~\cite{Hong2021VLNBERT} and DUET~\cite{Chen2022DUET}, have achieved strong benchmark performance on tasks such as R2R and RxR. More recent deployment-oriented work adopts video-VLM or VLA-style architectures and brings VLN closer to continuous embodied execution~\cite{Zhang2024NaVid,Wei2025StreamVLN,Zhu2026NaVIDA}. Although these advances improve policy capability, they do not by themselves resolve the runtime challenge of continuous execution. As a result, even policies with strong benchmark performance may still exhibit stop-and-go motion in streaming deployment.

\begin{figure}[!t]
\centering
\includegraphics[width=0.98\columnwidth]{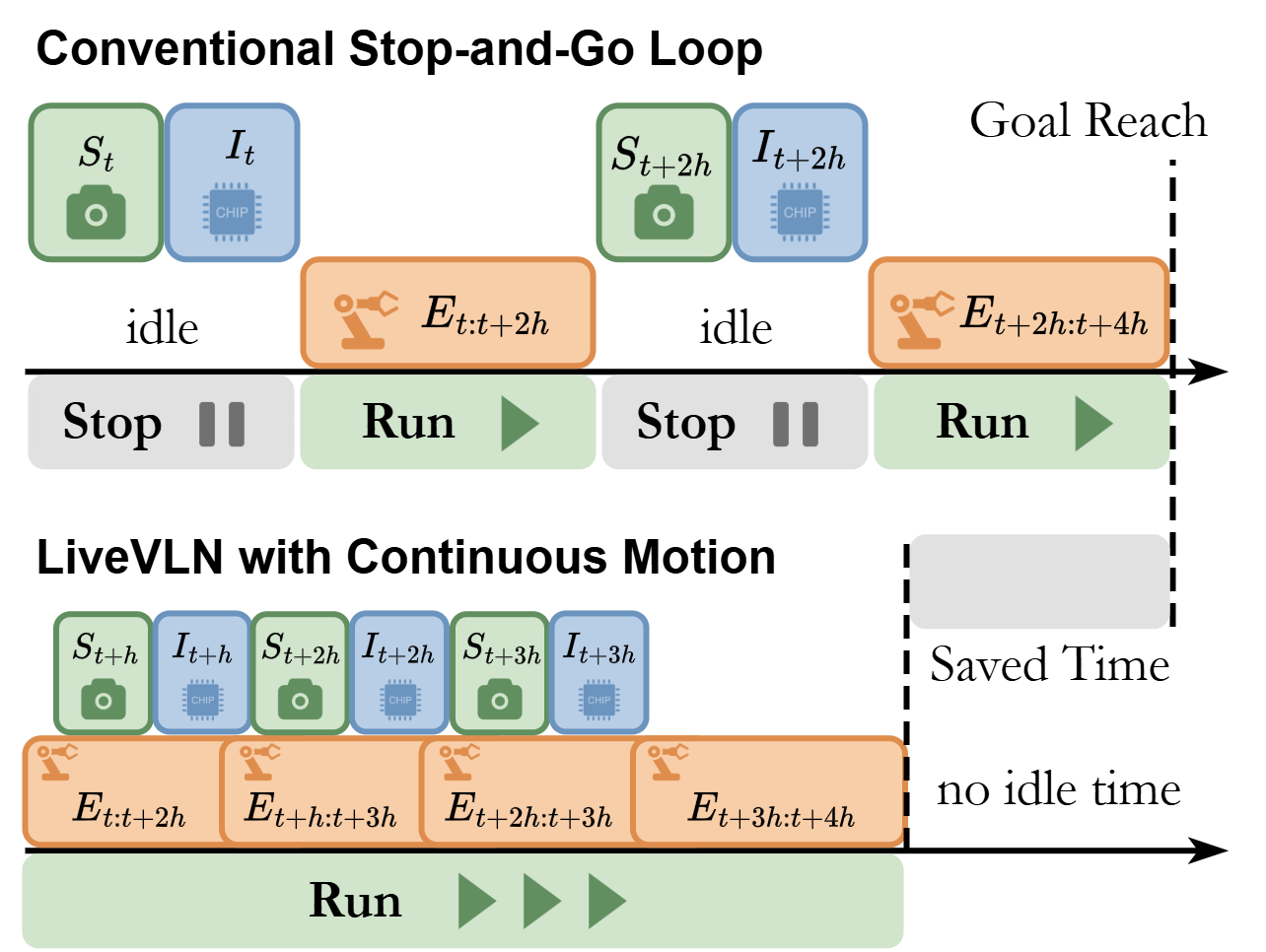}
\caption{Blocking VLN serializes \emph{sense(S)}, \emph{inference(I)}, and \emph{execution(E)}. After each action continuation, the robot pauses until the next sense-and-inference round finishes. \methodname\ instead keeps the execution thread running on a short guard buffer while the next sense-and-inference pass refreshes the following continuation in the background.}
\label{fig:teaser}
\end{figure}

The bottleneck is structural rather than purely computational. Most VLN systems still rely on a blocking three-stage interface consisting of {sense}, {inference}, and {execution}, in which the controller receives the next continuation only after sensing, transmission, preprocessing, and action generation on the latest observation are complete. If a continuation contains $k$ actions executed at frequency $f$, it supports only about $k/f$ seconds of motion, further reduced by controller-side execution overhead and communication jitter. Continuous execution therefore depends on whether the next continuation arrives before the current one is exhausted, exposing a direct trade-off between action freshness and motion continuity. Real-robot measurements show that this overhead is substantial. For example, in a native NaVIDA~\cite{Zhu2026NaVIDA} deployment, the controller waits for $10.64$\,s per episode on average, which accounts for a $30.5\%$ waiting ratio, and the system exhibits a pause count of $9.25$.

To address this issue, we introduce \methodname, a training-free runtime framework for continuous Vision-Language Navigation. The key idea is to decouple the current {execution} stage from the next {sense} and {inference} stages through a short-horizon action state with three roles: \emph{executed actions}, a released \emph{guard buffer} whose execution covers the next hidden sense-and-inference pass, and a \emph{revisable tail} that remains unreleased until handoff. One thread keeps executing the current guard buffer while the other thread refreshes the next continuation from the newest observation. \methodname\ combines two mechanisms: \emph{guarded handoff}, which switches to a refreshed continuation before the current guard buffer is depleted, and \emph{real-time adaptation}, under which the guard budget is determined by recent sense-and-inference latency (Fig.~\ref{fig:teaser}). In this way, only the minimal prefix necessary to sustain continuous motion is committed, leaving later actions open to revision based on newer observations. Operating at runtime, the framework applies to compatible pretrained VLM navigators that autoregressively decode multi-step action continuations~\cite{Zhang2024NaVid,Wei2025StreamVLN,Cheng2025NaVILA,Zhang2025UniNaVid,Zhu2026NaVIDA}.

We evaluate this runtime framework in Vision-Language Navigation in Continuous Environments (VLN-CE) benchmarks and native streaming deployment across multiple compatible pretrained VLM navigators. Beyond task success, we report continuity-oriented deployment metrics, including waiting time, waiting ratio, visible gap, pause count, and wall-clock episode duration, to examine whether the runtime hides sense-and-inference latency in practice. Under matched checkpoints, \methodname\ cuts waiting time by more than $70\%$ on both StreamVLN~\cite{Wei2025StreamVLN} and NaVIDA~\cite{Zhu2026NaVIDA}, shortens wall-clock episode time by $12.6$--$19.6\%$, and markedly reduces pause count, while preserving task performance in both simulation and real-robot trials.

Our contributions are as follows:
\begin{enumerate}
\item We analyze why blocking VLN deployment leads to stop-and-go motion, and we show through a real-robot NaVIDA diagnosis that the waiting ratio still reaches $30.5\%$ under the native runtime.
\item We introduce \methodname, a training-free runtime framework for compatible pretrained VLM navigators that organizes execution into executed actions, a guard buffer, and a revisable tail, enabling uninterrupted control through guarded handoff and real-time adaptation.
\item Across VLN-CE and native streaming deployment with StreamVLN and NaVIDA, \methodname\ reduces average waiting time by $77.7\%$ and $72.8\%$ and shortens wall-clock episode time by $12.6\%$ and $19.6\%$, respectively, while preserving benchmark performance.
\end{enumerate}

\section{Problem Formulation}
Blocking VLN deployment follows a serialized three-stage cycle of \emph{sense}, \emph{inference}, and \emph{execution}; real-robot measurements confirm that this interface induces stop-and-go behavior.

\subsection{Temporal Analysis of Blocking Deployment}

Given a language instruction $x$ and an online observation stream $\{o_r\}$, blocking deployment repeats three serialized stages in inference round $r$: \emph{sense} captures the newest observation, \emph{inference} maps it to the next action continuation, and \emph{execution} consumes the released actions on the controller. Let
\begin{equation}
b_r = [a_r^{(1)}, a_r^{(2)}, \dots, a_r^{(H_r)}],
\end{equation}
denote the continuation released after the round-$r$ sense-and-inference pass, where $H_r$ is its length and $a_r^{(i)}$ is its $i$-th action unit. If the controller has already issued or executed the first $p_r$ actions, with $p_r \in \{0,\dots,H_r\}$, then the remaining executable suffix is
\begin{equation}
b_r^{\mathrm{rem}} = [a_r^{(p_r+1)}, \dots, a_r^{(H_r)}].
\end{equation}
During the next round's sense-and-inference stage, this suffix is the controller's only executable buffer. Continuous motion requires the next continuation to arrive before the current execution buffer is exhausted; otherwise, the controller idles. Stop-and-go behavior reflects a timing mismatch between serialized sense-and-inference latency and the remaining execution budget.

For each episode, the wall-clock duration is decomposed as
\begin{equation}
\begin{aligned}
    T_{\mathrm{episode}} &= T_{\mathrm{execution}} + T_{\mathrm{wait}}, \\
    \eta_{\mathrm{wait}} &= \frac{T_{\mathrm{wait}}}{T_{\mathrm{episode}}},
\end{aligned}
\end{equation}
where $T_{\mathrm{execution}}$ is the cumulative \emph{execution} time and $T_{\mathrm{wait}}$ is the exposed \emph{sense} and \emph{inference} time during which no executable action is available. Each round is timestamped on the robot client at five events: sending the newest sensed observation $(t_{\mathrm{send}})$, receiving the server response $(t_{\mathrm{recv}})$, exhausting the current execution queue $(t_{\mathrm{empty}})$, issuing the first action from the refreshed queue $(t_{\mathrm{issue}})$, and finishing that control step $(t_{\mathrm{done}})$. These timestamps define the continuity metrics: inference latency $\ell_{\mathrm{infer}} = t_{\mathrm{recv}} - t_{\mathrm{send}}$, visible gap $\ell_{\mathrm{gap}} = t_{\mathrm{issue}} - t_{\mathrm{empty}}$, first resumed execution-step duration $\ell_{\mathrm{execution}} = t_{\mathrm{done}} - t_{\mathrm{issue}}$, and interrupted-round duration $\ell_{\mathrm{total}} = t_{\mathrm{done}} - t_{\mathrm{empty}} = \ell_{\mathrm{gap}} + \ell_{\mathrm{execution}}$. 
$N_{\mathrm{pause}}$ counts rounds with $\ell_{\mathrm{gap}} > 0.5\,\mathrm{s}$. Applying these definitions to the diagnostic study reveals a $30.5\%$ waiting ratio and frequent stop-and-go interruptions.

\begin{figure}[t]
\centering
\includegraphics[width=0.88\columnwidth]{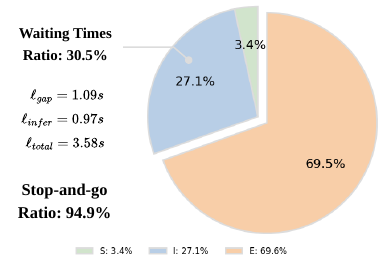}
\caption{Native blocking NaVIDA diagnosis on Unitree G1. 
The waiting ratio reaches $30.5\%$, while $94.9\%$ of inference rounds exhibit stop-and-go behavior.
}
\label{fig:findings_snapshot}
\end{figure}

\begin{figure*}[!t]
\centering
\includegraphics[width=0.88\textwidth]{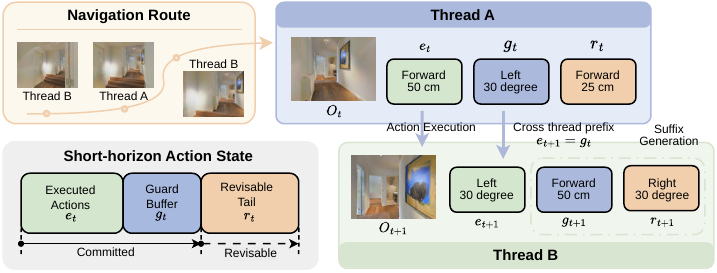}
\caption{Overview of \methodname\ when the next update arrives before the current guard buffer is exhausted. The short-horizon action state contains executed actions, a guard buffer, and a revisable tail. Thread A keeps the current guard buffer in \emph{execution}, while Thread B uses the newest sensed observation to run \emph{inference} for the next continuation. At handoff, the current guard buffer becomes executed actions, the refreshed prefix becomes the next guard buffer, and the trailing tail remains revisable for correction after the next observation arrives.}
\label{fig:framework}
\end{figure*}

\subsection{Real-Robot Diagnostic Study}
The diagnostic study uses a controlled NaVIDA deployment on a Unitree G1 robot with a front-facing Intel RealSense D455f camera and monocular RGB input. The robot performs \emph{sense} and low-level \emph{execution}, while a remote server with an NVIDIA RTX 5090 GPU performs high-level \emph{inference} and returns the next action continuation through the motion-control API. Episode-level waiting and per-round discontinuity are measured on this system.

\paragraph{Episode-Level Evidence}
Average waiting is $10.64$\,s per episode, corresponding to a $30.5\%$ waiting ratio. The controller experiences $9.25$ pauses over $9.75$ inference rounds on average, so $94.9\%$ of rounds are stop-and-go. Native blocking NaVIDA spends nearly one third of a typical episode in waiting time instead of continuous execution.

\paragraph{Per-Round Discontinuity Evidence}
After the current queue is exhausted, the controller waits $1.09$\,s on average before motion resumes. This visible gap is close to the $0.97$\,s inference round trip, so most of the sense-and-inference pass remains visible to the robot. Once the refreshed action arrives, the first resumed execution step lasts $2.49$\,s on average. Native blocking deployment therefore produces frequent, roughly one-second stalls. StreamVLN exhibits the same qualitative pattern under the same deployment protocol (Table~\ref{tab:continuity_main_results}).

These results motivate a dual-thread runtime that overlaps sense and inference with execution while preserving a short-horizon action state composed of executed actions, a guard buffer, and a revisable tail for online revision.

\section{Method}

\methodname\ mitigates stop-and-go deployment without changing the pretrained VLN policy. It replaces the serialized \emph{sense}-\emph{inference}-\emph{execution} interface with a dual-thread runtime that lets execution continue during the next sense-and-inference pass. At each round, the controller keeps executing the current guard buffer while a background inference thread refreshes a new continuation in parallel; if the refresh finishes before the guard is exhausted, handoff moves that guard into executed actions, promotes the refreshed prefix to the next guard buffer, and leaves the remaining tail revisable. Given an autoregressive navigator $\mathcal{B}$ that emits multi-step executable continuations, the runtime uses two coupled mechanisms. \emph{Guarded handoff with a revisable tail} organizes each short-horizon action state into executed actions, a guard buffer, and a revisable tail, while \emph{real-time adaptation} sizes that guard buffer against deployment latency across compatible pretrained VLM navigators. \methodname\ treats $\mathcal{B}$ as a continuation generator conditioned on the current guard buffer and defines the corresponding handoff and release rules.

At inference round $t$, \emph{sense} provides the newest observation, \emph{inference} predicts the next continuation, and \emph{execution} consumes the current guard buffer on the controller. Relative to Sec.~II, the continuation $b_r$ is now written as $u_t$ and split into executed actions $e_t$, guard buffer $g_t$, and revisable tail $r_t$. Let $c_t$ denote the full conditioning context at inference round $t$ (instruction, current observation, and any model-specific history input), and let $u_t=[a_t^{(1)},\dots,a_t^{(H_t)}]$ denote the continuation predicted by $\mathcal{B}$. The runtime never exposes all of $u_t$ at once: it releases only the committed prefix needed for continuity and keeps the remaining tail revisable until the next handoff. The next refresh is therefore conditioned on the already committed guard buffer $g_t=[a_t^{(1)},\dots,a_t^{(k)}]$ rather than on the still-revisable tail. The backbone's native autoregressive factorization is unchanged; only the runtime handoff boundary exposed to the controller moves.

\subsection{Guarded Handoff with a Revisable Tail}\label{subsec:dual-thread}
\methodname\ targets compatible pretrained VLM navigators that infer multi-step action continuations autoregressively during online execution. The navigator must (i) take the latest sensed observation and the already committed guard buffer as inference inputs, and (ii) emit an ordered multi-step sequence of executable action units, where each unit may be a primitive action or a controller-accepted macro-action. Recent VLM-style VLN systems satisfy these assumptions and already use multi-step action continuations~\cite{Zhang2024NaVid,Wei2025StreamVLN,Cheng2025NaVILA,Zhang2025UniNaVid,Zhu2026NaVIDA}. The handoff rule depends only on the emitted continuation, not on backbone-specific architectural design or retraining.

In streaming deployment, the instruction remains fixed throughout the episode and visual observations arrive sequentially during motion. The runtime controller maintains the following short-horizon interface state
\begin{equation}
q_t=[e_t \mid g_t \mid r_t],
\end{equation}
which decomposes the current short-horizon state into three runtime roles: executed actions, a guard buffer, and a revisable tail.

Equivalently, the same continuation is viewed as already executed actions, a currently executable guard buffer, and a still-hidden revisable tail.

\paragraph{Executed Actions}
$e_t$ records the actions that have been executed in the current short-horizon state. It excludes earlier episode history and tracks only what has been consumed since the latest handoff.

\paragraph{Guard Buffer}
$g_t$ is the released guard buffer currently held by the controller. It provides the continuity budget that keeps motion ongoing while the next hidden sense-and-inference pass runs in the background.

\paragraph{Revisable Tail}
$r_t$ is the unreleased tail still held by the background inference thread. Because it has not been released to the controller, it can be replaced by a refreshed inference result from newer observations. The boundary between $g_t$ and $r_t$ separates committed actions from revisable actions.

At round $t$, Thread A keeps executing the released guard buffer $g_t$ without interruption, while Thread B uses the newest sensed observation $o_{t+1}$ to refresh the continuation for the next handoff. The next handoff succeeds only if this Thread-B sense-and-inference pass finishes before the current guard buffer is exhausted. Thread B computes the refreshed continuation as
\begin{equation}
\hat{u}_{t+1}=\mathcal{D}_{\mathcal{B}}(o_{t+1}, g_t ; r_t),
\end{equation}
where $\mathcal{D}_{\mathcal{B}}$ denotes the backbone-specific continuation generator. The length $H_{t+1}$ may be fixed or round-dependent. Here $g_t$ is the committed context that must survive the handoff, whereas $r_t$ is the revisable tail and may be overwritten by the refresh. The released guard buffer $g_t$ keeps the controller in execution while the background inference thread completes the next sense-and-inference pass.

At handoff, the current guard buffer is fully executed, so $e_{t+1}=g_t$ and the refreshed continuation is split into $q_{t+1}=[e_{t+1} \mid g_{t+1} \mid r_{t+1}]$. The guard buffer in round $t$ becomes executed actions in round $t+1$, the refreshed prefix becomes the next guard buffer, and the remaining tail stays revisable until the following refresh.

\begin{algorithm}[t]
\caption{Core pipeline of \methodname}
\label{alg:handoff}
\small
\begin{algorithmic}[1]
\Require $q_t=[e_t \mid g_t \mid r_t]$, $o_{t+1}$, $\tilde{\ell}_{t,\mathrm{SI}}$, $\alpha$, $\delta$, and $T_t^{\mathrm{guard}}$
\Ensure candidate split $\hat{u}_{t+1}=[\hat{g}_{t+1}\mid \hat{r}_{t+1}]$
\Statex \textcolor{blue!70!black}{$\triangleright$ \textbf{[Sec.~III-A] Background refresh}}
\State Keep executing $g_t$ while running $\hat{u}_{t+1} \gets \mathcal{D}_{\mathcal{B}}(o_{t+1}, g_t ; r_t)$
\State Measure the resulting sense-and-inference latency $\ell_{t,\mathrm{SI}}$
\Statex \textcolor{blue!70!black}{$\triangleright$ \textbf{[Sec.~III-B] Real-time adaptation}}
\State $\tilde{\ell}_{t+1,\mathrm{SI}} \gets (1-\alpha)\tilde{\ell}_{t,\mathrm{SI}} + \alpha \ell_{t,\mathrm{SI}}$
\State $\psi_{t+1} \gets \tilde{\ell}_{t+1,\mathrm{SI}} + \delta$
\State Choose $k_{t+1}^{\star} = \min \{k : T_{t+1}^{(k)} \ge \psi_{t+1}\}$
\State Split $\hat{u}_{t+1}$ into $\hat{g}_{t+1}$ and $\hat{r}_{t+1}=\hat{u}_{t+1}^{(>k_{t+1}^{\star})}$
\Statex \textcolor{blue!70!black}{$\triangleright$ \textbf{[Sec.~III-B] Release and backup action}}
\State Check whether $\ell_{t,\mathrm{SI}} \le T_t^{\mathrm{guard}}$
\State If true, release $\hat{g}_{t+1}$ and keep $\hat{r}_{t+1}$ revisable; otherwise trigger a backup action or \stopaction
\end{algorithmic}
\end{algorithm}

\subsection{Real-Time Adaptation and Transferability}
\label{subsec:adaptive-guard}

\begin{table*}[t]
\caption{RGB-only VLN-CE results on R2R and RxR \texttt{val\_unseen}. Metrics include navigation error (NE), oracle success (OS), success rate (SR), success weighted by path length (SPL), and normalized dynamic time warping (nDTW). Rows marked ``+ \methodname'' use the same pretrained checkpoint with our training-free runtime. \textsuperscript{$\dagger$} denotes extra navigation data, and \texttt{--} means unreported.}
\label{tab:task_main_results}
\centering
{\footnotesize
\setlength{\tabcolsep}{3pt}
\renewcommand{\arraystretch}{1.02}
\begin{tabular}{lcccccccc}
\toprule
Method & \multicolumn{4}{c}{R2R \texttt{val\_unseen}} & \multicolumn{4}{c}{RxR \texttt{val\_unseen}} \\
\cmidrule(lr){2-5}\cmidrule(lr){6-9}
& NE $\downarrow$ & OS $\uparrow$ & SR $\uparrow$ & SPL $\uparrow$ & NE $\downarrow$ & SR $\uparrow$ & SPL $\uparrow$ & nDTW $\uparrow$ \\
\midrule
NaVid \cite{Zhang2024NaVid} & 5.47 & 49.1 & 37.4 & 35.9 & -- & -- & -- & -- \\
MapNav \cite{Zhang2025MapNav} & 4.93 & 53.0 & 39.7 & 37.2 & -- & -- & -- & -- \\
NaViLA \cite{Cheng2025NaVILA} & 5.37 & 57.6 & 49.7 & 45.5 & -- & -- & -- & -- \\
NaViLA\textsuperscript{$\dagger$} \cite{Cheng2025NaVILA} & 5.22 & 62.5 & 54.0 & 49.0 & 6.77 & 49.3 & 44.0 & 58.8 \\
UniNaVid\textsuperscript{$\dagger$} \cite{Zhang2025UniNaVid} & 5.58 & 53.3 & 47.0 & 42.7 & 6.24 & 48.7 & 40.9 & -- \\
\addlinespace[2pt]
StreamVLN \cite{Wei2025StreamVLN} & 5.43 & 62.5 & 52.8 & 47.2 & 6.72 & 48.6 & 42.5 & 60.2 \\
StreamVLN\textsuperscript{$\dagger$} \cite{Wei2025StreamVLN} & 4.90 & 63.6 & 56.4 & 50.2 & 5.65 & 54.4 & 45.4 & 63.7 \\
\rowcolor{gray!15}
\hspace{0.8em}+ \methodname\textsuperscript{$\dagger$} (training-free) & 4.84 & 64.4 &  57.2 & 50.0 & 5.48 &  53.7 & 44.5 & 63.0 \\
NaVIDA \cite{Zhu2026NaVIDA} & 4.32 & 69.5 & 61.4 & 54.7 & 5.23 & 57.4 & 49.6 & 67.0 \\
\rowcolor{gray!15}
\hspace{0.8em}+ \methodname\ (training-free) & 4.51 & 67.8 & 59.9 & 53.9 & 5.22 & 56.7 & 49.2 & 67.3 \\
\bottomrule
\end{tabular}
}
\end{table*}

The controller must retain executable actions while the next sense-and-inference pass runs in the background. Because primitive actions and macro-actions can occupy the controller for very different durations, the guard is sized in wall-clock time rather than in action counts. The scheduler therefore seeks the shortest released prefix whose predicted execution can cover the next hidden refresh.

Given the refreshed continuation $\hat{u}_{t+1}$, each action unit $a$ has an estimated controller-side execution time $\tau(a)$. The predicted total execution time for the first $k$ actions in $\hat{u}_{t+1}$ is
\begin{equation}
T_{t+1}^{(k)} = \sum_{i=1}^{k} \tau(\hat{a}_{t+1}^{(i)}).
\end{equation}
If the next hidden sense-and-inference pass is expected to require a time budget $\psi_{t+1}$, the next guard is the shortest prefix whose cumulative execution time can cover that budget:
\begin{equation}
k_{t+1}^{\star}
=
\min
\left\{
k \in \{1,\dots,H_{t+1}\}
\mid
T_{t+1}^{(k)} \ge \psi_{t+1}
\right\},
\end{equation}
with $k_{t+1}^{\star}=H_{t+1}$ if no shorter prefix satisfies the condition. The resulting split is $\hat{u}_{t+1}=[\hat{g}_{t+1}\mid \hat{r}_{t+1}]$, where $\hat{g}_{t+1}$ is the next guard buffer and $\hat{r}_{t+1}$ is the revisable tail. Only the shortest future prefix needed to preserve continuity is exposed, while the remaining tail stays available for correction from newer observations.

Real-time adaptation comes from updating $\psi_{t+1}$ according to recent sense-and-inference latency. After round $t$, we update a running latency estimate and the next guard budget as
\begin{equation}
\tilde{\ell}_{t+1,\mathrm{SI}}
=(1-\alpha)\tilde{\ell}_{t,\mathrm{SI}}+\alpha \ell_{t,\mathrm{SI}},
\qquad
\psi_{t+1}=\tilde{\ell}_{t+1,\mathrm{SI}}+\delta,
\end{equation}
where $\alpha \in (0,1]$ is the update factor and $\delta \ge 0$ is a safety margin. The exponential average follows latency drift while smoothing single-round noise.

The adaptive scheduler sizes the round-$(t+1)$ guard buffer for the \emph{next} hidden sense-and-inference pass. For the \emph{current} round, continuity depends on whether the ongoing background inference finishes before the already released guard buffer is exhausted. Let $\ell_{t,\mathrm{SI}}$ be the latency of the current background sense-and-inference pass and let $T_t^{\mathrm{guard}}$ be the predicted execution time of the currently released guard buffer. Seamless handoff requires
\begin{equation}
\ell_{t,\mathrm{SI}} \le T_t^{\mathrm{guard}}.
\end{equation}
When the condition holds, the system releases $\hat{g}_{t+1}$ and retains $\hat{r}_{t+1}$ as the revisable tail. If the background pass misses the guard deadline but a previously validated one-step backup action still covers the overrun, the controller emits that backup action; otherwise it emits \stopaction. Continuity is therefore adjusted online rather than by a fixed offline horizon.

The same rule transfers across compatible pretrained VLM navigators and robot platforms. The runtime estimates the next sense-and-inference latency, releases the shortest prefix as the next guard buffer whose execution time covers that delay, and keeps the remainder as the revisable tail using only continuation output and action-timing estimates. No backbone-specific retraining, architectural change, or fixed action horizon is required beyond the interface in Sec.~\ref{subsec:dual-thread}. Moving to another compatible navigator or embodiment mainly requires re-estimating $\tau(\cdot)$ and measuring the local sense-and-inference latency; the guarded-handoff and adaptive-release logic remains unchanged.

\section{Experiments}\label{sec:experiments}

\begin{table*}[t]
\caption{Real-robot continuity and efficiency for native StreamVLN and NaVIDA versus their \methodname\ wrappers under matched robot, scene, and checkpoints. Up arrows indicate higher-is-better metrics, and down arrows indicate lower-is-better metrics.}
\label{tab:continuity_main_results}
\centering
{\scriptsize
\setlength{\tabcolsep}{3pt}
\renewcommand{\arraystretch}{1.05}
\begin{tabular}{lccccccccccc}
\toprule
Method & \multicolumn{7}{c}{\textit{Episode-level}} & \multicolumn{4}{c}{\textit{Per-round timing}} \\
\cmidrule(lr){2-8}\cmidrule(lr){9-12}
& $\bar{N}_{\mathrm{round}}$ & $\bar{T}_{\mathrm{execution}}$ & $\bar{T}_{\mathrm{wait}} \downarrow$ & $\bar{T}_{\mathrm{episode}} \downarrow$ & $\bar{\eta}_{\mathrm{wait}} \downarrow$ & $\bar{N}_{\mathrm{pause}} \downarrow$ & Stop within $2$\,m $\uparrow$ & $\bar{\ell}_{\mathrm{infer}} \downarrow$ & $\bar{\ell}_{\mathrm{gap}} \downarrow$ & $\bar{\ell}_{\mathrm{execution}}$ & $\bar{\ell}_{\mathrm{total}} \downarrow$ \\
\midrule
StreamVLN & 7.60 & 34.66 & 7.32 & 41.98 & 17.4\% & 6.75 & 15/40 & 0.81 & 0.96 & 4.56 & 5.52 \\
\rowcolor{gray!15}
StreamVLN + \methodname & 14.80 & 35.08 & 1.63 & 36.71 & 4.4\% & 0.80 & 16/40 & 0.52 & 0.11 & 2.37 & 2.48 \\
NaVIDA & 9.75 & 24.26 & 10.64 & 34.90 & 30.5\% & 9.25 & 12/40 &  0.97 & 1.09 & 2.49 & 3.58 \\
\rowcolor{gray!15}
NaVIDA + \methodname & 18.25 & 25.17 & 2.89 & 28.06 & 10.3\% & 1.20 & 12/40 & 0.61 & 0.16 & 1.38 & 1.54 \\
\bottomrule
\end{tabular}
}
\end{table*}

We evaluate \methodname\ from three perspectives: simulation task performance, deployment continuity under native real-robot streaming, and ablations. Simulation results are reported on R2R and RxR \texttt{val\_unseen}. Real-robot results are measured with a shared client-side timing protocol and cover continuity, wall-clock efficiency, and whether the final stopping position falls within $2$\,m of the ground-truth goal.

\subsection{Experimental Setup}
We evaluate VLN-CE on R2R and RxR \texttt{val\_unseen} under the same streaming interface. For StreamVLN and NaVIDA \cite{Wei2025StreamVLN,Zhu2026NaVIDA}, all comparisons use the same checkpoints and deployment settings as their native runtimes. Table~\ref{tab:task_main_results} reports the standard R2R/RxR metrics \cite{Anderson2018R2R,Ku2020RxR}: navigation error (NE), oracle success (OS), success rate (SR), success weighted by path length (SPL), and normalized dynamic time warping (nDTW), which respectively measure endpoint accuracy, whether the path ever reaches the goal region, final success, success under path-efficiency weighting, and trajectory-path fidelity. For StreamVLN, both the offline and real-robot results use the stronger extra-navigation-data checkpoint.

For the real-robot study, both navigators are deployed on the same Unitree G1 client-server platform in a shared scene, with Wi-Fi jitter remaining within $50$\,ms during collection. Each navigator is evaluated for $40$ runs ($8$ instructions $\times$ $5$ repetitions). Table~\ref{tab:continuity_main_results} reports episode- and round-level continuity metrics together with wall-clock episode time and whether the robot's final stopping position falls within $2$\,m of the ground-truth goal.

Because the two navigators expose different native action primitives and sense-and-inference cadences, we keep each method in its intended runtime configuration. For StreamVLN \cite{Wei2025StreamVLN}, the native runtime decodes and executes a 4-action continuation per round, whereas \methodname\ releases only a short guard buffer and updates the revisable tail online; $\tau(a)$ is therefore the fixed action duration implied by the control frequency. On the real robot, considering this inference-execution timing pattern and StreamVLN's reuse of KV-cache states across multi-turn decoding, we fix the guard buffer to two actions and place those two actions in the execution buffer at each handoff. For NaVIDA, each action unit is a native controller chunk rather than a fixed-duration primitive, so \methodname\ commits the shortest prefix whose predicted execution covers the next hidden sense-and-inference budget, with $\tau(a)$ estimated from empirical controller-side durations.

\subsection{Main Results}

Table~\ref{tab:task_main_results} shows that \methodname\ preserves benchmark navigation quality while changing only the runtime framework. Across both StreamVLN and NaVIDA, the wrapped runtime stays close to the native checkpoint, with only small and mixed fluctuations on R2R and RxR. This pattern indicates that the gains below come from the runtime framework rather than from retraining or stronger benchmark scores.

Table~\ref{tab:continuity_main_results} shows a much larger effect in real-robot streaming. Across both navigators, \methodname\ reduces waiting time by more than $70\%$, markedly reduces pause count, and shortens wall-clock episode time from $41.98$\,s to $36.71$\,s on StreamVLN and from $34.90$\,s to $28.06$\,s on NaVIDA. Meanwhile, success under the ``stop within $2$\,m of the ground-truth goal'' criterion remains comparable, indicating that the main gain comes from the runtime framework rather than from a change in endpoint capability.

The per-round timing statistics support the same interpretation. With \methodname, the visible gap drops from about one second to about one tenth of a second, while the number of inference rounds increases substantially. For StreamVLN, the real-robot numbers also explain why the fixed two-action guard is the right operating point: one atomic action lasts about $1.14$\,s on average, so a two-action guard covers about $2.28$\,s of controller time, whereas the hidden inference round takes only $0.52$\,s on average. A single-action guard would leave much less slack once sensing, communication jitter, and handoff overhead are included, while a longer fixed guard would unnecessarily shrink the revisable tail and weaken online correction. In other words, more frequent refresh becomes useful only when it is hidden behind committed execution instead of being exposed as repeated blocking stalls.

\begin{figure}[!t]
\centering
\includegraphics[width=0.95\columnwidth]{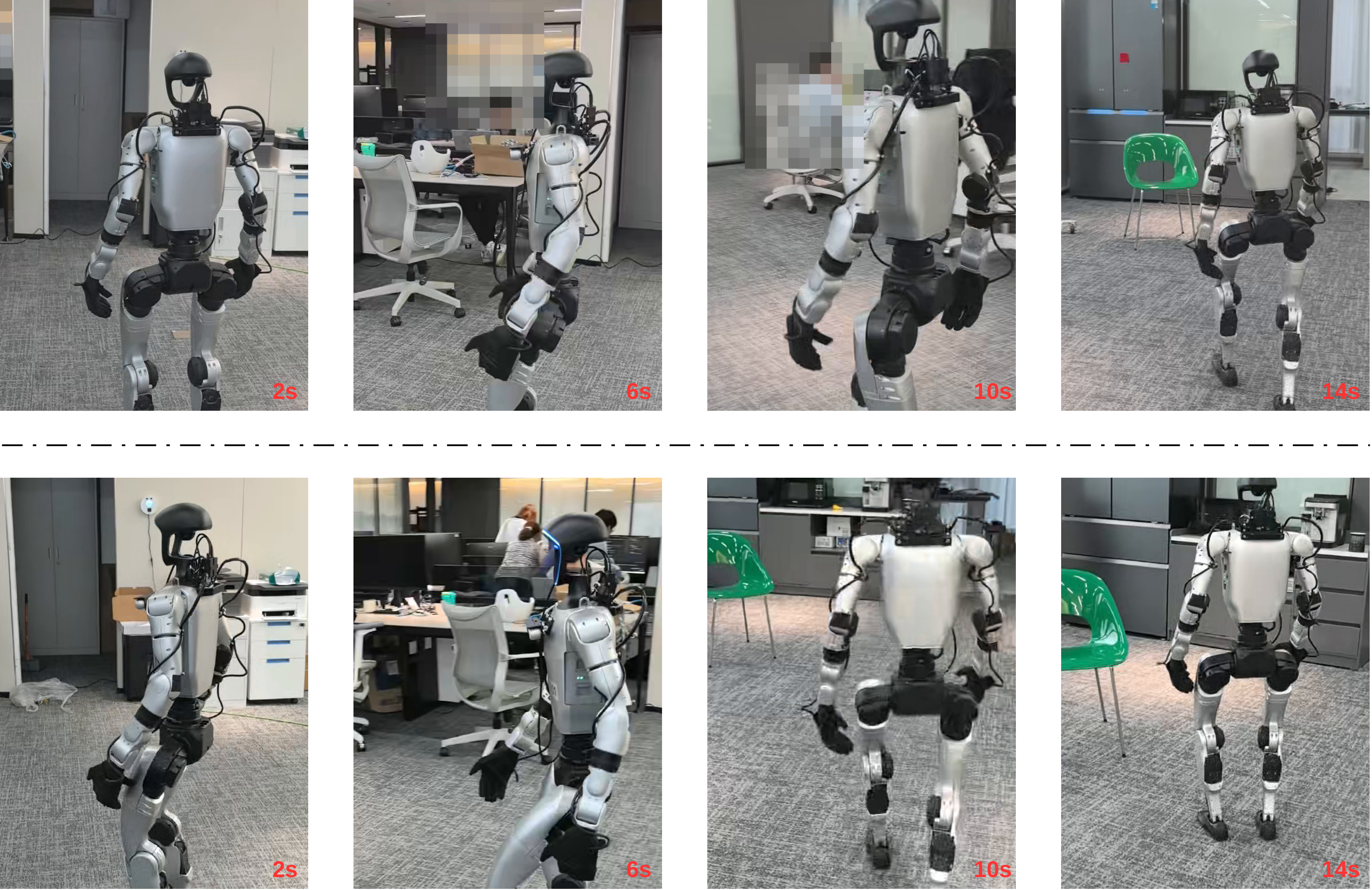}
\caption{Real-robot deployment in the shared office scene for the instruction ``turn left until you see the green chair, walk to the green chair, and stop beside it.'' Top: native blocking StreamVLN. Bottom: \methodname\ under the same setup. \methodname\ moves more continuously and completes the task in 14\,s, while native StreamVLN remains visibly farther from the goal under the same wall-clock budget overall.}
\label{fig:real_world}
\end{figure}

Figure~\ref{fig:real_world} qualitatively corroborates Table~\ref{tab:continuity_main_results}. Under the same instruction and wall-clock budget, \methodname\ maintains sustained motion and reaches the goal region, whereas the native blocking StreamVLN runtime still exhibits stop-and-go behavior and remains visibly farther from the target.

\subsection{Ablation Study}

\begin{table}[t]
\caption{NaVIDA ablation on simulation and real-robot continuity. $\bar{N}_{\mathrm{round}}$ denotes average inference rounds per episode.}
\label{tab:navida_ablation_plan}
\centering
{\footnotesize
\setlength{\tabcolsep}{1.5pt}
\renewcommand{\arraystretch}{1.1}
\begin{tabular}{@{}p{0.43\columnwidth}cccc@{}}
\toprule
Variant &  R2R SR $\uparrow$ & $\bar{N}_{\mathrm{round}}$ & $\bar{\eta}_{\mathrm{wait}} \downarrow$ & $\bar{N}_{\mathrm{pause}} \downarrow$ \\
\midrule
Native blocking reference &  61.4 &9.75 & 30.5\% & 9.25 \\
Blocking w/ more rounds &  61.0 &18.50 & 52.2\% & 17.75 \\
\midrule
Full \methodname &  59.9 &18.25 & 10.3\% & 1.20 \\
\methodname\ w/o revisable tail &  57.6 &16.50 & 15.6\% & 2.25 \\
\shortstack[l]{\methodname\ w/o real-time adaptation} & 60.3 &  19.75 & 18.4\% & 3.00 \\
\bottomrule
\end{tabular}
}
\end{table}

Table~\ref{tab:navida_ablation_plan} focuses on inference rounds, waiting ratio, and pause count, with R2R \texttt{val\_unseen} SR included as a stability reference. The first comparison shows that increasing refresh frequency alone does not solve the deployment problem. Relative to the native blocking reference, \textit{Blocking w/ more rounds} nearly doubles the average inference frequency from $9.75$ to $18.50$, yet the waiting ratio rises from $30.5\%$ to $52.2\%$. By contrast, full \methodname\ runs at a similarly high inference frequency while reducing the waiting ratio to $10.3\%$ and the pause count to $1.20$. The key difference is therefore not how often the system refreshes, but whether refresh is overlapped with execution so that the updated prefix is handed off before the committed actions are exhausted.

The component ablations further clarify which part of the design provides which benefit. Removing the revisable tail weakens both task stability and continuity, indicating that the hidden tail remains important after part of the continuation has already been committed. Removing real-time adaptation shows a different pattern: SR remains comparable, but the waiting ratio rises to $18.4\%$ and the pause count to $3.00$. Taken together, these results suggest that the revisable tail primarily preserves adaptability, whereas real-time adaptation primarily preserves continuity by matching the released guard budget to the latency of the next hidden refresh. This is also why the fixed StreamVLN setting keeps exactly two guard actions: it is the smallest practical committed prefix that safely exceeds the measured hidden refresh budget on the robot, whereas NaVIDA uses the same rule in wall-clock form by selecting the shortest controller chunk prefix whose predicted execution covers the next inference pass.

\subsection{Discussion}
\paragraph{Real-Robot Continuity Depends Strongly on How Runtime Latency Is Exposed}
The main difference between the native runtime and \methodname\ is not policy quality but how sense-and-inference latency is presented to the controller. When that latency is exposed directly, the robot experiences visible stop-and-go motion; when it is hidden behind committed execution, motion becomes substantially more continuous without materially changing benchmark performance.

\paragraph{Asynchronous Overlap Alone Is Not Enough}
The ablation study shows that the gain does not come from issuing more refreshes by itself. Full \methodname\ achieves the best continuity, whereas removing either the revisable tail or real-time adaptation weakens performance. The two components also play different roles: the revisable tail contributes more to preserving task success, while real-time adaptation contributes more directly to hiding latency. Continuity therefore improves only when refresh is paired with a revisable tail and a real-time guard budget, so that new evidence can still update the tail without re-exposing the controller to a blocking stall.

\paragraph{Efficiency Gains Remain Bounded By Physical Execution}
Reducing waiting time also shortens episode time, but the effect is limited since physical execution still dominates. This bounded episode-time gain is expected from the decomposition above: \methodname\ reduces $T_{\mathrm{wait}}$ without speeding up the physical execution component $T_{\mathrm{execution}}$. This explains why the continuity gain is large while the episode-time gain is more moderate.

\paragraph{Implication For Deployment-Oriented VLN Evaluation}
Taken together, deployment-oriented VLN should evaluate not only endpoint success but also whether actions remain continuously available at execution time. Two systems with similar benchmark SR can produce noticeably different robot behavior if one exhibits a higher pause count. Our findings therefore support treating continuity-related timing measures as first-class deployment metrics alongside standard navigation benchmarks such as SR and SPL.

\section{Related Work}
\subsection{Vision-Language Navigation}
VLN began with instruction-following benchmarks such as R2R and RxR \cite{Anderson2018R2R,Ku2020RxR}, and early progress focused on improving next-action prediction through stronger grounding, large-scale pretraining, and richer history modeling \cite{Wang2019RCM,Hao2020PREVALENT,Hong2021VLNBERT,Chen2021HAMT}. Subsequent work introduced more explicit spatial structure and longer-horizon reasoning, for example through topological planning, dual-scale graph reasoning, and map-centric representations \cite{Chen2021TopoPlanning,Chen2022DUET,Huang2022VLMaps,An2025ETPNav}. As VLN evolved from discrete viewpoints to continuous environments, the problem also became more operational: low-level embodiment exposes control drift, recovery difficulty, and timing effects that are only weakly reflected in stepwise evaluation \cite{Krantz2020VLNCE,Hong2022Bridge}. More recent scaling and LLM-assisted approaches further expanded the design space by increasing training diversity and explicit reasoning capacity \cite{Wang2023ScaleVLN,Zhou2024NavGPT}.

Recent work has moved VLN closer to online embodied deployment. NaVid uses a video-based VLM for continuous instruction following \cite{Zhang2024NaVid}, while StreamVLN improves online action prediction through streaming context management, including a multi-rate context design and KV-cache reuse \cite{Wei2025StreamVLN}. NaVILA further couples a navigation-oriented VLA with a real-time locomotion policy for legged robots \cite{Cheng2025NaVILA}. NaVIDA strengthens action-grounded visual dynamics with inverse-dynamics augmentation and adaptive action-continuation execution \cite{Zhu2026NaVIDA}, and JanusVLN improves RGB-only VLN with a decoupled semantic-spatial design that supports efficient incremental updates \cite{Zeng2025JanusVLN}. On the training side, ActiveVLN studies active exploration through multi-turn reinforcement learning for VLN post-training \cite{Zhang2025ActiveVLN}, and SACA introduces step-aware dense supervision to improve credit assignment in VLN-CE optimization \cite{Li2026SACA}. These efforts move VLN closer to online deployment, but they mainly strengthen policy capability, representation quality, or training signals. A separate question is how such policies should be executed under latency, streaming observations, and controller-side safety constraints. Our work targets that runtime level through a training-free runtime framework rather than proposing another policy architecture or post-training method.

\subsection{Streaming VLMs and Live Video Understanding}
Broader multimodal modeling has recently shifted from offline video understanding toward streaming and live interaction. Recent work spans long-video modeling for preserving temporal context over pre-collected videos \cite{Zhang2025LLaVAVideo,Qian2024VideoStreaming}, native streaming systems that update context and respond continuously as frames arrive \cite{Chen2024VideoLLMOnline,Liu2025StreamChat,Wang2025StreamBridge}, and training or evaluation efforts that better expose live multimodal behavior, such as large-scale streaming supervision in LiveCC and streaming-video benchmarks such as StreamingBench \cite{Chen2025LiveCC,Lin2024StreamingBench}. Parallel real-time inference has also been identified as a core requirement for genuine live understanding rather than post-hoc response generation \cite{Lin2026SpeakWhileWatching}. Together, these studies suggest that live multimodal systems require persistent state, incremental updates, and non-blocking inference.

Embodied navigation shares these requirements, but adds a control constraint: actions must remain executable while execution itself changes future observations. The setting therefore lies between streaming multimodal inference and receding-horizon control, where partial action continuations can be exposed early and refreshed as observations evolve \cite{Leviathan2023Speculative,Mayne2000MPC}. From this perspective, \methodname\ is closer to a receding-horizon execution interface than to live video captioning, because it commits a short safe prefix, refreshes the remainder asynchronously, and replans as observations evolve.

\section{Conclusion}
We presented \methodname, a training-free runtime framework that keeps executable actions available while refreshing future continuations online. After integration with compatible pretrained VLM navigators, benchmark performance on R2R and RxR remains close to the native runtime, while real-robot deployment improves substantially: waiting time drops by more than $70\%$, pause count is greatly reduced, and wall-clock episode time is shortened. Taken together, these results suggest that stop-and-go VLN behavior is not only a policy-quality issue but also a runtime issue. They also motivate treating continuity and navigation efficiency as first-class deployment objectives alongside standard navigation benchmarks such as SR and SPL, rather than as secondary runtime details.

\bibliographystyle{IEEEtran}
\bibliography{refs}

\end{document}